\documentclass[12pt]{article}	
\usepackage[margin=20mm]{geometry}
\usepackage{graphicx}
\usepackage{amsmath}
\usepackage{amsthm}
\usepackage{here, latexsym, amssymb, bm, ascmac, mathtools, multicol, tcolorbox, subfig}
\usepackage{listings}
\usepackage{xcolor}
\lstdefinelanguage{Stan}{
  keywordstyle=\color{blue},
  morekeywords={data, parameters, model, real, vector, matrix, int, transformed, parameters, generated},
  sensitive=false,
  morecomment=[l]{//},
  morecomment=[s]{/*}{*/},
  morestring=[b]",
}

\newtheorem{teiri}{Theorem}

\newtheorem{rei}{Example}

\usepackage{enumerate}
\usepackage{ascmac,fancybox}
\usepackage{abstract}
\theoremstyle{plain}
\definecolor{darkblue}{rgb}{0.00,0.00,0.55}
\definecolor{darkslateblue}{rgb}{0.28,0.24,0.55}
\definecolor{darkgreen}{rgb}{0.00,0.39,0.00}

\definecolor{Brown}{cmyk}{0,0.81,1,0.60}
\definecolor{OliveGreen}{cmyk}{0.64,0,0.95,0.40}
\definecolor{CadetBlue}{cmyk}{0.62,0.57,0.23,0}

\title{Learning under Singularity: \\An Information Criterion improving WBIC and sBIC}
  
\author{Lirui Liu and Joe Suzuki\\Osaka University}
\date{}

\begin{document}
\maketitle

\begin{abstract}
We introduce a novel Information Criterion (IC), termed Learning under Singularity (LS), designed to enhance the functionality of the Widely Applicable Bayes Information Criterion (WBIC) and the Singular Bayesian Information Criterion (sBIC). LS is effective without regularity constraints and demonstrates stability. Watanabe defined a statistical model or a learning machine as regular if the mapping from a parameter to a probability distribution is one-to-one and its Fisher information matrix is positive definite. In contrast, models not meeting these conditions are termed singular. Over the past decade, several information criteria for singular cases have been proposed, including WBIC and sBIC. WBIC is applicable in non-regular scenarios but faces challenges with large sample sizes and redundant estimation of known learning coefficients. Conversely, sBIC is limited in its broader application due to its dependence on maximum likelihood estimates. LS addresses these limitations by enhancing the utility of both WBIC and sBIC. It incorporates the empirical loss from the Widely Applicable Information Criterion (WAIC) to represent the goodness of fit to the statistical model, along with a penalty term similar to that of sBIC. This approach offers a flexible and robust method for model selection, free from regularity constraints.\\

\end{abstract}



\section{Introduction}

We introduce a novel information criterion to address the limitations inherent in WBIC and sBIC, enhancing stability and applicability without imposing regularity constraints. A statistical model or learning machine is deemed regular if the mapping from parameters to probability distributions is one-to-one and if its Fisher information matrix is always positive definite (Watanabe, 2010 \cite{watanabe2}). Conversely, it is labeled singular. Many statistical models and learning machines exhibit singularity rather than regularity; any model featuring hierarchical layers, latent variables, or grammatical rules is considered singular. Existing information criteria such as AIC (Akaike Information Criterion) and BIC (Bayesian Information Criterion) assume the true distribution is regular with respect to the statistical model. In models characterized by regularity, the Bayes free energy, derived as the negative logarithm of the Bayes marginal likelihood, can be asymptotically estimated using BIC. However, in the case of a singular statistical model where the likelihood function defies approximation by a normal distribution, both AIC and BIC become impractical for model assessment. Consequently, constructing singular learning theory is crucial in both statistics and learning theory.

Watanabe (2013) \cite{watanabe2} defined a widely applicable Bayesian information criterion (WBIC) as the average log-likelihood function over the posterior distribution with an inverse temperature $\beta > 0$. They demonstrated that WBIC shares the same asymptotic expansion as the Bayes free energy, even for singular or unrealizable statistical models. WBIC can be numerically computed without knowledge of the true distribution, so it represents a generalized version of BIC for singular statistical models.

However, WBIC encounters challenges with large sample sizes and redundant estimation of known learning coefficients. Moreover, its outcome depends on selecting $\beta_0$, with different choices potentially yielding conflicting results.

Another generalized version of BIC for singular cases was proposed by Drton et al. (2017) \cite{sbic}, known as the Singular Bayesian Information Criterion (sBIC). 
Reduced rank regression, factor analysis, and latent class analysis represent singular submodels of an exponential family, whether normal or multinomial. 
Consequently, the sequence of likelihood ratios 
$p(x_i|\hat{\theta}_n)/p(x_i|\theta)$
converges in distribution and is bounded in probability (Drton, 2009 \cite{Drton3}). In such cases, $\theta_n$ is substituted with the maximum likelihood estimator $\hat{\theta}_n$. 
Building upon asymptotic theory for the marginal likelihood of singular models (Watanabe, 2009 \cite{watanabe3}), sBIC was introduced in the form:
\begin{eqnarray*}
sBIC=\sum_{i=1}^{n}-\log p\left(x_{i} \mid\hat{\theta}_n\right)+\lambda \log n.
\end{eqnarray*}
For more complex models like Gaussian mixture models, likelihood ratios can often be proven to converge in distribution under compactness assumptions on the parameter space. sBIC captures the large sample behavior of the marginal likelihood integrals through theoretical insights into the learning coefficient $\lambda$ (Drton et al., 2017 \cite{sbic}).

While sBIC indeed extends BIC to non-regular cases, it still has constraints. For instance, assumption 1 from Section 4 of their paper may fail in some scenarios, such as mixture model setups with unbounded parameter spaces (Hartigan, 1985 \cite{Hartigan}).

Moreover, sBIC is valid only if the parameter space is compact. Even if this condition holds, the maximum likelihood estimator $\hat{\theta}$ may sometimes be unobtainable, thereby limiting sBIC's performance in general cases and specific scenarios. Specifically, akin to AIC and BIC, sBIC still utilizes the quantity:
\begin{eqnarray*}
\sum_{i=1}^n\left\{-\log p\left(x_i \mid \hat{\theta}_n\right)\right\}
\end{eqnarray*}
to gauge the goodness of fit to the statistical model. In a singular case where the Hessian matrix is non-invertible, numerical optimization algorithms like the Newton–Raphson method may fail to converge to a specific estimate. It's worth noting that if the maximum likelihood does not exist, neither does the maximum likelihood estimation nor the values of the information criterion derived from it.

In essence, sBIC estimates the true likelihood using the maximum likelihood estimator to extend Schwarz's BIC to regular models. Alternative estimation methods, such as posterior means as used by Roeder and Wasserman (1997) \cite{Roeder}, can also prove effective.

One advantage of sBIC over WBIC is its independence from Markov chain Monte Carlo (MCMC) approximation of the posterior distribution. However, it relies on theoretical results regarding learning coefficients; only a few are currently known. To overcome this challenge, Imai and Kuroki \cite{Imai1} (see also Yoshioka and Suzuki, 2018 \cite{yoshioka}) propose using WBIC to derive a consistent estimator $\hat{\lambda}$ of $\lambda$, then plugging $\hat{\lambda}$ into sBIC, thereby potentially widening its applicability. However, this approach sacrifices the computational efficiency advantage of sBIC.

To overcome the limitations of WBIC and sBIC, we introduce a novel Information Criterion, Learning under Singularity (LS), which enhances stability and applicability without regularity constraints. While WBIC struggles with large sample sizes and redundant estimation of known learning coefficients, we enhance its applicability by incorporating a penalty term similar to sBIC. We establish that LS demonstrates increased stability with larger sample sizes through mathematical proofs. Unlike WBIC, whose outcomes vary based on the choice of $\beta_0$, leading to potential confusion, LS yields a single clear result. On the other hand, sBIC's performance is not universally guaranteed, especially in cases where obtaining the maximum likelihood estimator is challenging, such as when the Hessian matrix is non-invertible. To address this challenge, we utilize the empirical loss from WAIC to assess the goodness of fit to the statistical model, thereby departing from the constraints of maximum likelihood estimation. LS may appear less efficient computationally than sBIC because it requires MCMC, akin to WBIC. However, when learning coefficients are unknown, this difference diminishes, as sBIC necessitates WBIC to estimate learning coefficients and still relies on MCMC.

The remainder of this paper is structured as follows: Section 2 provides the necessary background to comprehend the results presented herein. Section 3 introduces the information criterion (LS) and derives its theoretical properties. Section 4 showcases two experimental results. Finally, Section 5 concludes this paper and proposes avenues for future research.


\section{Preliminaries}
\subsection{Regularity}
Let $X$ and $\mathcal{X}$ be a random variable and its sample space, with the true distribution denoted by $q$.
We seek a parameter $\theta\in \Theta \subseteq \mathbb{R}^{d}$, where $d\geq 1$, such that 
the Kullback-Leibler (KL) divergence
\begin{eqnarray*}
D(q \| p) := \mathbb{E}_{X}\left[\log \frac{q(X)}{p(X \mid \theta)}\right]=\int_{\mathcal{X}} q(x) \log \frac{q(x)}{p(x \mid \theta)} \, dx \geq 0
\end{eqnarray*}
is minimized for a statistical model $\{p(\cdot \mid \theta)\}_{\theta \in \Theta}$. 
Let $\Theta_{*}$ denote the set of all $\theta$ that minimize $D(q \| p)$. For each pair $\theta_*\in\Theta_*$ and $\theta\in \Theta$, we define the quantity
\begin{eqnarray*}
K(\theta) :=
\mathbb{E}_{X}\left[\log \frac{p\left(X \mid \theta_{*}\right)}{p(X \mid \theta)}\right]=
D(q \| p)-\mathbb{E}_{X}\left[\log \frac{q(X)}{p\left(X \mid \theta_{*}\right)}\right].
\end{eqnarray*}
We immediately find the equivalence
$$D(q\|p)\ {\rm is\ minimized}\ \Longleftrightarrow K(\theta)=0 \Longleftrightarrow \theta\in \Theta_*\ .$$
According to the theory of Watanabe \cite{WAIC, Suzuki23}, the model $\{p(\cdot|\theta)\}_{\theta\in \Theta}$ is said to be regular with respect to $q$ (we write ``$(p,q)$ are regular'') if and only if the following conditions are satisfied:
\begin{enumerate}[(i)]
    \item There exists a unique $\theta_{*} \in \Theta$ such that $\Theta_{*}=\left\{\theta_{*}\right\}$.
    \item The Hessian $\left.\nabla^{2}K(\theta)\right|_{\theta=\theta_{*}}$ is positive definite.
    \item The optimal parameter $\theta_{*}$ is not on the boundary of $\Theta$.
\end{enumerate}
Let $\nabla K(\theta)=\left(\displaystyle \frac{\partial K(\theta)}{\partial \theta_i}\right)
$ and $J(\theta):=\nabla^2 K(\theta)=
\left(\displaystyle \frac{\partial^2 K(\theta)}{\partial \theta_i\partial \theta_j}\right)
$ 
be the vector and matrix 
obtained by differentiating $K(\theta)$ by
$\theta_1,\ldots,\theta_d$ once and twice, respectively. Then, 
regularity implies that exactly $d$ second-order variables exist
in $K(\theta)$:
\begin{eqnarray*}
K(\theta)&=&K(\theta_*)+(\theta-\theta_*)\nabla K(\theta)|_{\theta=\theta_*}+
\frac{1}{2}(\theta-\theta_*)^\top \nabla^2 K(\theta)|_{\theta=\theta_*}(\theta-\theta_*)+\cdots\nonumber\\
&=&\frac{1}{2}(\theta-\theta_*)^\top J(\theta_*)(\theta-\theta_*)+ (power\ series\ of\ {\theta}_1,\ldots,{\theta}_d) \label{eq0-2211}\\
&=&\frac{1}{2}(\lambda_1\tilde{\theta}_1^2+\cdots+\lambda_d\tilde{\theta}_d^2)+ (power\ series\ of\ \tilde{\theta}_1,\ldots,\tilde{\theta}_d)\label{eq0-2212}
\end{eqnarray*}
for transformed variables $\tilde{\theta}_1,\ldots,\tilde{\theta}_d$,
where $\lambda_1,\ldots,\lambda_d>0$ are the $d$ eigenvalues of $J(\theta)$.

\subsection{Learning Coefficient}
We define the learning coefficient, denoted by $\lambda$, with respect to the prior distribution $\varphi(\cdot)$ and the kernel function $K(\cdot)$ for a true parameter value $\theta_* \in \Theta_*$. 
This coefficient is determined by finding the largest pole of the zeta function:
\begin{eqnarray*}
\zeta(z) = \int_\Theta K(\theta)^z \varphi(\theta), d\theta,
\end{eqnarray*}
where $z$ is the complex variable, the order $m$ of the pole is defined as its multiplicity. It is established that all poles of $\zeta(z)$ are real and negative numbers \cite{watanabe23}. For regular cases where the prior distribution is positive and the distributions $(p,q)$ are regular, it is known that $\lambda \leq d/2$, where $d$ is the dimension of the parameter space.

\begin{rei}\rm
Consider the case where $p(\cdot|\theta)$ and $q(\cdot)$ follow normal distributions $N(\theta,1)$ and $N(0,1)$ respectively, and the prior distribution $\varphi(\theta)$ is given by:
\begin{eqnarray*}
\varphi(\theta) =
\begin{cases}
\frac{1}{2}, & \text{if } |\theta| \leq 1,\\
0, & \text{otherwise}.
\end{cases}
\end{eqnarray*}
In this scenario, the kernel function is given by 
\begin{eqnarray*}
K(\theta)=\int_{-\infty}^\infty \frac{1}{\sqrt{2\pi}}\exp(-\frac{x^2}{2})\left\{\frac{(x-\theta)^2}{2}-\frac{x^2}{2}\right\}dx=\frac{\theta^2}{2}\ ,
\end{eqnarray*}
leading to the zeta function:
\begin{eqnarray*}
\zeta(z) = \int_{-1}^1 \left(\frac{\theta^2}{2}\right)^z \frac{1}{2} d\theta = \frac{1}{2^z(2z+1)}\frac{1-(-1)^{2z+1}}{2},
\end{eqnarray*}
resulting in $\lambda=1/2$.
\end{rei}

In cases involving singularities, such as those encountered in algebraic geometry, learning coefficients must be derived using the blow-up method. However, the identification of such coefficients remains limited, with only a few examples, including the formula by Aoyagi \cite{Aoyagi} used in the experiments of Section 4 (Table \ref{ao}), having been derived thus far.

For problems where the values of learning coefficients are unknown, methods for estimating them from actual data are available. However, the approximation error is significant, especially when dealing with small sample sizes (Section 3).

\subsection{sBIC and WBIC}
Suppose we have obtained $n$ i.i.d. samples $x_1,\ldots,x_n\in {\cal X}$. 
Let $\hat{\theta}_n$ be 
the maximum-likelihood estimation that 
maximizes the likelihood $\prod_{i=1}^n p\left(x_i \mid \theta\right)$, or minimizes the negative log-likelihood
\begin{eqnarray*}
l:=-\frac{1}{n} \sum_{i=1}^n \log p\left(x_i \mid \theta\right)
\end{eqnarray*}
over $\theta\in \Theta$.
Assume that the learning coefficient $\lambda$ has been obtained somehow. 
The Singular Bayesian Information Criterion(Drton et al., 2017\cite{sbic}) is defined by
\begin{eqnarray*}
sBIC:=\sum_{i=1}^{n}-\log p\left(x_{i} \mid \hat{\theta}_n\right)+\lambda \log n\ .
\end{eqnarray*}

Assume that $(p,q)$ are regular. When $\Theta$ is compact, 
the estimator $\hat{\theta}_n$ converges in probability to $\theta_{*}$ as $n \rightarrow \infty$ (Suzuki, 2023\cite{Suzuki23}). 
Note that the performance of sBIC is not guaranteed in all singular cases. In fact, the maximum-likelihood estimator $\hat{\theta}$ sometimes may not be obtained, and sBIC can not be represented using it. For example, if the Hessian matrix is non-invertible, then numerical optimization algorithms e.g. Newton–Raphson method that solves
$\nabla l =0$ and iterates 
$$z_{i+1}=z_i -(\nabla^2 l)^{-1}\nabla l$$
for $i=1,2,\ldots$ from some $z_1\in \Theta$
may fail to converge because $\nabla^2 l\approx \nabla^2 K(\theta)$ that has no inverse, which is attributed to the difficulties in determining the direction of parameter updates during the optimization process, making it challenging to find parameter values that maximize the likelihood function.

Watanabe(2013)\cite{watanabe2} introduces a parameter $\beta>0$, called the inverse temperature, and assumes that it decreases with $n$ as $\beta=\beta_0/\log n$ for some constant $\beta_0>0$. Given $x_1,\ldots,x_n\in \mathcal{X}$, define the posterior mean of a function $f:\Theta \rightarrow \mathbb{R}$ with respect to $\beta$ by
\begin{equation}\label{eq1}
\mathcal{E}_{\beta}[f(\theta)]:=
\frac
{\displaystyle \int_{\Theta} f(\theta) \prod_{i=1}^n p(x_i|\theta)^\beta \varphi(\theta) d\theta}
{\displaystyle \int_{\Theta} \prod_{i=1}^n p(x_i|\theta')^\beta \varphi(\theta') d\theta'}
\end{equation} 
The Watanabe Bayesian Information Criterion(Watanabe, 2013\cite{watanabe2}) is defined as
\begin{eqnarray*}
WBIC_n:=\mathcal{E}_{\beta}\left[\sum_{i=1}^{n}-\log p\left(x_{i} \mid \theta\right)\right].
\end{eqnarray*}
From (Watanabe, 2013\cite{watanabe2}), WBIC behaves as
\begin{align}\label{eq2}
WBIC_{n}=\sum_{i=1}^{n}-\log p\left(x_{i} \mid \theta_{*}\right)+\frac{\lambda \log n}{\beta_{0}}+U_{n} \sqrt{\frac{\lambda \log n}{2 \beta_{0}}}+O_{p}(1),
\end{align}
where $U_{1}, U_{2}, \ldots$ are random variables that converge in law to a normal distribution with mean zero.

WBIC also gives a method to estimate the learning coefficient if a true distribution is unknown. However, if theoretical results about it are known, the estimation becomes redundant. 

\section{Proposed Information Criterion: Learning under Singularity (LS)}

We first define the predictive distribution $r(x|x_1,\ldots,x_n)$ and the empirical loss $T_n$ as follows:
\begin{eqnarray*}
r(x | x_{1}, \ldots, x_{n}) = \mathcal{E}1[p(x|\theta)], \quad x\in\mathcal{X},
\end{eqnarray*}
and
\begin{eqnarray*}
T{n} = \frac{1}{n} \sum_{i=1}^{n}\left\{-\log r(x_{i} | x_{1}, \ldots, x_{n})\right\},
\end{eqnarray*}
respectively.

We propose a novel information criterion, Learning under Singularity (LS), defined by:
\begin{equation}\label{eq:LS}
\text{LS} = n T_{n} + \lambda \log n.
\end{equation}

If the value of $\lambda$ is unknown, it can be estimated using $\hat{\lambda}$, obtained from:
\begin{align}\label{eq:lambda_hat}
\hat{\lambda} = \frac{WBIC_{1} - WBIC_{2}}{1 / \beta_{1} - 1 / \beta_{2}} = \lambda + O_{P}\left(\frac{1}{\sqrt{\log n}}\right),
\end{align}
where this formula is derived from equation (\ref{eq:WBIC}).

In comparison with WBIC and sBIC, LS offers several advantages. Unlike sBIC, which relies on maximum-likelihood estimates, LS employs empirical loss.

Our superiority over WBIC is supported by the following theorem:
\begin{teiri}
\begin{align}
\text{LS} &= \sum_{i=1}^{n}\left\{-\log p\left(x_{i} | \theta_{}\right)\right\} + \lambda \log n + O_{P}(1). \label{eq:LS_expanded} \\
&= \sum_{i=1}^{n}\left\{-\log p\left(x_{i} | \theta_{}\right)\right\} + \hat{\lambda} \log n + O_{P}(\sqrt{\log n}). \label{eq:LS_with_hat_lambda}
\end{align}
\end{teiri}
\textbf{Proof.} According to Suzuki (2023), $T_{n}$ can be expressed as:
\begin{eqnarray*}
T_{n} = \frac{1}{n} \sum_{i=1}^{n}\left\{-\log p\left(x_{i} | \theta_{}\right)\right\} + \frac{1}{n}\left(\lambda - \frac{1}{2} \mathcal{E}\left[\sqrt{t} \xi_{n}(u) | x_{1}, \ldots, x_{n}\right]\right)
-\frac{1}{2} \mathbb{E}{X}[\mathcal{V}(X)] + o_{P}\left(\frac{1}{n}\right),
\end{eqnarray*}
where $\mathcal{E}[\cdot]$ and $\mathcal{V}[\cdot]$ are the posterior expectation and variance in equation (\ref{eq1}) with $\beta=1$, respectively. Since $\lambda$ is constant and $n \mathbb{E}{X}[\mathcal{V}(X)]$ and $\mathcal{E}\left[\sqrt{t} \xi{n}(u) | x_{1}, \ldots, x_{n}\right]$ converge to their expected values by the law of large numbers, they are $O_{P}(1)$. Substituting them into the definition of LS (\ref{eq:LS}), we obtain (\ref{eq:LS_expanded}). Substituting $\lambda$ with $\hat{\lambda}$, LS becomes (\ref{eq:LS_with_hat_lambda}). This completes the proof.

Regarding WBIC, assuming $\beta_0=1$, equation (\ref{eq:WBIC}) becomes:
\begin{equation}\label{eq:WBIC}
WBIC_{n} = \sum_{i=1}^{n}-\log p\left(x_{i} | \theta_{*}\right) + \lambda \log n + O_{P}(\sqrt{\log n}),
\end{equation}
where, in equation (\ref{eq:WBIC}), the third term divided by $\sqrt{\log n}$ is $O_P(1)$.

Based on equations (\ref{eq:LS_expanded}), (\ref{eq:LS_with_hat_lambda}), and (\ref{eq:WBIC}), we conclude:
\begin{enumerate}[(I)]
\item If $\lambda$ is unknown, LS and WBIC perform similarly as $n \rightarrow \infty$.
\item If $\lambda$ is known, LS outperforms WBIC.
\end{enumerate}
The first statement is justified because equation (\ref{eq:LS_with_hat_lambda}) approximates equation (\ref{eq:WBIC}) with a bounded, negligible quantity and a constant. Regarding the second statement, equation (\ref{eq:WBIC}) contains a term $O_{P}(\sqrt{\log n})$, which becomes unstable for large $n$.

\section{Experiments} 

\subsection{Application to Reduced-Rank Regression Models}
Reduced-rank regression is a variant of multivariate regression where the rank of the linear transformation between inputs and outputs is constrained to be smaller than the dimensions of the inputs and outputs. 
This implies a reduction in the number of hidden variables compared to a standard multivariate regression model. 
Consider a three-layer neural network with $M$ units in the input layer, $H$ units in the hidden layer, and $N$ units in the output layer. 
Assuming the model is realizable, 
we denote the parameter matrices as $A \in \mathbb{R}^{H \times M}$ and $B \in \mathbb{R}^{N \times H}$, with their true values denoted as $A_*$ and $B_*$. 
The relationship between input $x$ and output $y$ can be expressed as $y - B_* A_* x \sim N(0, I_N)$, where $I_N$ is the identity matrix of size $N$. Let $\varphi(\theta_) > 0$ represent the prior distribution at the true parameters, and let $r:= \text{rank}(B_* A_*)$. When the true rank $r$ satisfies $r \leq \min{H, M, N}$, learning coefficients (Aoyagi \cite{Aoyagi}) as shown in Table \ref{ao} can be derived.

\begin{table}
\caption{Aoyagi Formula for Reduced Rank Regression \cite{Aoyagi}}\label{ao}
{\small
\begin{center}
\begin{tabular}{c|ccc|c|c|c}
\hline & $M+r$ & $N+r$ & $H+r$ & $M+H$ & & \\
Case & $:$ & $:$ & $:$ & + & $m$ & $\lambda$ \\
& $N+H$ & $M+H$ & $M+N$ & $N+r$ & & \\
\hline $1 \mathrm{a}$ & $\leq$ & $\leq$ & $\leq$ & even & 1 & $-(H+r-M-N)^2 / 8+M N / 2$ \\
\cline { 5 - 7 } $1\mathrm{b}$ & & & & odd & 2 & $-(H+r-M-N)^2 / 8+M N / 2+1 / 8$ \\
\hline 2 & $>$ & & & & 1 & $(H N-H r+M r) / 2$ \\
\hline 3 & & $>$ & & & 1 & $(H M-H r+N r) / 2$ \\
\hline 4 & & & $>$ & & 1 & $M N / 2$ \\
\hline
\end{tabular}
\end{center}
}
\end{table}
For instance, consider $M=3$, $N=4$, $H=3$, and $r=3$. Using Aoyagi's formula and \eqref{eq:lambda_hat}, we find that $\lambda$ and $\hat{\lambda}$ are $6$ and $6.329$, respectively. We observe that the values of $\lambda$ and $\hat{\lambda}$ are close. Here, we utilize $\hat{\lambda}$ to calculate LS. Comparing WBIC and LS, we find that they exhibit similar performances, with LS consistently slightly larger than WBIC by a constant, supporting statement (I). It is noteworthy that WBIC varies with $\beta_0$, while LS remains independent of it.

\begin{table}[H]
\setlength{\abovecaptionskip}{0cm}
		\setlength{\belowcaptionskip}{-0.2cm}
\caption{\small Computing WBIC and LS for reduced-rank regression models, $M=3, N=4, H=3, r=3$}\label{tab1}
\begin{center}
\begin{tabular}{|l|l|l|}
\hline & WBIC & LS \\
\hline$\beta_{0}=3$ & 212.63048450177914 & 224.42567234 \\
\hline$\beta_{0}=5$ & 211.50724201207544 & 224.42567234 \\
\hline$\beta_{0}=7$ & 211.00680119632437 & 224.42567234 \\
\hline$\beta_{0}=10$ & 210.65029139809852 & 224.42567234 \\
\hline
\end{tabular}
\end{center}
\end{table}

\subsection{Application to Gaussian Mixture Models}
Consider a Gaussian mixture model with $H$ Gaussian distributions, each chosen randomly with probabilities $\pi_{1}, \ldots, \pi_{H} \geq 0$ such that $\sum_{h=1}^{H} \pi_{h}=1$. Data $x \in \mathbb{R}^N$ of $N$ dimensions is generated according to these Gaussian distributions $N\left(\mu_h, \sigma_h^2\right)$. The Gaussian mixture model is defined as
$$
p(x \mid \theta)=\sum_{h=1}^H \pi_h s_h\left(x \mid \mu_h,\sigma_h^2\right)
$$
where $\theta=\left\{\left(\pi_h, \mu_h\right) \mid h=1, \ldots, H\right\}$ and $s_h\left(x \mid \mu_h, \sigma_h^2\right)$ is the probability density function of the normal distribution with mean $\mu_h$ and variance $\sigma_h^2$.

Assuming the above statistical model is realizable, we denote the true value of $H$ as $H_*$ and the true parameters as
$$
\theta_*=\left\{\left(\pi_h^*, \mu_h^*\right) \mid h=1, \ldots, H_*\right\}
$$

For Gaussian mixture models, Suzuki \cite{Suzuki23} provides an upper bound for $\lambda$: $\displaystyle \lambda \leq \frac{1}{2}\left(N H_{*}+H-1\right)$. In the following examples, we calculate the values of WBIC and LS for observed values $x_1, \ldots, x_n \in \mathbb{R}^N$ and candidate values of $H$.

In the first example, with $\mu=(1,1)$ and the identity matrix of size $N=2$ as $\Sigma$, we generate $n=200$ data points each from $N(-\mu, \Sigma)$ and $N(\mu, \Sigma)$. We calculate WBIC and LS for $H=1,2,3,4$, considering $\pi_{1}^{}=\pi_{2}^{}=1 / 2$ and $H_{*}=2$. WBIC estimates correctly when $\beta_{0}=1$, but not when $\beta_{0}=10$, while LS's result is correct.

Similarly, we generate $n=600$ data points each from three Gaussian distributions $N(-2\mu, \Sigma)$, $N(0, \Sigma)$, and $N(2\mu, \Sigma)$. We calculate the values of Gaussian mixture WBIC and LS for $H=1,2,3,4$, considering $\pi_{1}^{}=\pi_{2}^{}=\pi_{3}^{}=1 / 3$ and $H_{}=3$. Again, WBIC's performance is unstable compared to LS.

\begin{table}[H]
\setlength{\abovecaptionskip}{0cm}
		\setlength{\belowcaptionskip}{-0.2cm}
\begin{center}
\caption{\small Computing WBIC and LS for $n=200, N=2, H_{*}=2, H=1,2,3,4$}
\begin{tabular}{|c|c|c|c|c|}
\hline \multicolumn{5}{|l|}{WBIC}  \\
\hline & $H=1$ & $H=2$ & $H=3$ & $H=4$ \\
\hline$\beta_{0}=1$ & \begin{tabular}{l}
770.699654036304
\end{tabular} & \begin{tabular}{l}
\color{red}688.176545308359
\end{tabular} & \begin{tabular}{l}
688.415656722353
\end{tabular} & \begin{tabular}{l}
688.823035296964
\end{tabular} \\
\hline$\beta_{0}=10$ & \begin{tabular}{l}
766.656827569317
\end{tabular} & \begin{tabular}{l}
677.675220262638
\end{tabular} & \begin{tabular}{l}
\color{red}675.41674081187
\end{tabular} & \begin{tabular}{l}
675.591697101393
\end{tabular} \\
\hline \multicolumn{5}{|l|}{LS}  \\
\hline & $H=1$ & $H=2$ & $H=3$ & $H=4$ \\
\hline  & \begin{tabular}{l}
391.11623602499003
\end{tabular} & \begin{tabular}{l}
\color{red}349.8558642044485
\end{tabular} & \begin{tabular}{l}
351.4600822779085
\end{tabular} & \begin{tabular}{l}
353.953978268919
\end{tabular} \\
\hline
\end{tabular}
\end{center}
\end{table}

\begin{table}[H]
\setlength{\abovecaptionskip}{0cm}
		\setlength{\belowcaptionskip}{-0.2cm}
\begin{center}
\caption{\small Computing WBIC and LS for $n=600, N=3, H_{*}=3, H=1,2,3,4$}
\begin{tabular}{|c|c|c|c|c|}
\hline \multicolumn{5}{|l|}{WBIC}  \\
\hline & $H=1$ & $H=2$ & $H=3$ & $H=4$ \\
\hline$\beta_{0}=1$ & \begin{tabular}{l}
3280.76865311864
\end{tabular} & \begin{tabular}{l}
2331.54286502827
\end{tabular} & \begin{tabular}{l}
\color{red}2247.19376372556
\end{tabular} & \begin{tabular}{l}
2248.16537133019
\end{tabular} \\
\hline$\beta_{0}=10$ & \begin{tabular}{l}
3276.16268953823
\end{tabular} & \begin{tabular}{l}
2319.77837797933
\end{tabular} & \begin{tabular}{l}
2226.53921853956
\end{tabular} & \begin{tabular}{l}
\color{red}2225.67320347581
\end{tabular} \\
\hline \multicolumn{5}{|l|}{LS}  \\
\hline & $H=1$ & $H=2$ & $H=3$ & $H=4$ \\
\hline  & \begin{tabular}{l}
1105.34397244879
\end{tabular} & \begin{tabular}{l}
789.9424869301365
\end{tabular} & \begin{tabular}{l}
\color{red}762.989514258491
\end{tabular} & \begin{tabular}{l}
765.391060701189
\end{tabular} \\
\hline
\end{tabular}
\end{center}
\end{table}
\section{Conclusion}
Our paper introduces the innovative information criterion, Learning under Singularity (LS), for model selection. Distinguishing itself from sBIC, LS operates effectively beyond the restriction of maximum likelihood, making it applicable in a wider range of singular scenarios. When compared with WBIC, LS demonstrates enhanced performance, especially in cases where the learning coefficient is known.

One future challenge lies in comparing the values with the free energy. While WBIC has an arbitrary parameter $\beta_0$, the proposed LS does not require the selection of $\beta_0$. It would be desirable to demonstrate that, even without selecting $\beta_0$, LS can yield values close to the free energy.

\appendix
\section{Stan code}
\vspace{-0.1cm}
\subsection{Stan code for Application to Reduced-Rank Regression Models}
\vspace{-0.1cm}

\begin{lstlisting}
functions {
  real ld_diag_lpdf ( real x , int i , int k , real c0 ) {
    return (k -i )* log (x) - square ( x) /(2* c0 );
  }
}
data {
  int n;
  int M;
  int N;
  int H;
  matrix [n , M ] X ;
  matrix [n , N ] Y ;
  real < lower =0 > beta ;
}
transformed data {
  int ntrap = M*H;
}
parameters {
  real < lower =0 > sigma ;
  real < lower =0 > nu ;
  vector < lower =0 >[ H] diags ;
  vector [ ntrap ] lowtrap ;
  matrix [H , N ] BhatT ;
  vector < lower =0 >[ M] lambda ;
}
transformed parameters {
  matrix [n , N ] mu ;
  matrix [M , H ] L ;
  {
  int idx ;
  idx =0;
  L = rep_matrix (0 , M , H);
  for ( col in 1: H ) {
    L[ col , col ] = diags [ col ];
    for (r in ( col +1) : M) {
      idx +=1;
      L[r , col ] = lowtrap [ idx ];
    }
  }
  mu = diag_post_multiply (X , lambda )*L* BhatT ;
  }
}
model {
  lowtrap ~ normal (0 ,1) ;
  for (i in 1: H)
    diags [ i] ~ ld_diag (i , H , 1) ;
  to_vector ( BhatT ) ~ normal (0 ,1) ;
  lambda ~ normal (0 , 2) ;
  nu ~ gamma (2 , 0.1) ;
  sigma ~ cauchy (0 ,1) ;
  for (j in 1: n ){
    for (i in 1: N )
      target += beta * student_t_lpdf (Y[j ,i ]| nu , mu [j ,i] , sigma );
  }
}
generated quantities { 
  matrix [n ,N] log_lik ;
  for (j in 1: n) {
    for (i in 1: N )
      log_lik [j ,i ] = student_t_lpdf (Y[j ,i ]| nu , mu [j ,i], sigma ) ;
  }
}
\end{lstlisting}

\begin{lstlisting}
data {
  int<lower=1> K;          
  int<lower=1> N;          
  vector[2] y[N];               
  real beta;
}
parameters {
  simplex[K] theta;          
  vector[2] mu[K];   
}
transformed parameters{
  vector[K] log_theta = log(theta);  
}
model {
  mu ~ multi_normal(rep_vector(0.0,2), diag_matrix(rep_vector(1.0,2)));
  for (n in 1:N) {
    vector[K] lps = log_theta;
    for (k in 1:K)
      lps[k] += multi_normal_lpdf(y[n] | mu[k], diag_matrix(rep_vector(1.0,2)));
    target += beta*log_sum_exp(lps);
  }
}
generated quantities{
  vector[N] log_lik;
  for (n in 1:N) {
    vector[K] lps = log_theta;
    for (k in 1:K)
      lps[k] += multi_normal_lpdf(y[n] | mu[k], diag_matrix(rep_vector(1.0,2)));
    log_lik[n] = log_sum_exp(lps);
  }
}
\end{lstlisting}

\vspace{-0.5cm}
\subsection{Stan code for Application to Gaussian Mixture Models}
\vspace{-0.1cm}

\begin{lstlisting}
data {
  int<lower=1> K;          
  int<lower=1> N;          
  vector[2] y[N];               
  real beta;
}
parameters {
  simplex[K] theta;          
  vector[2] mu[K];   
}
transformed parameters{
  vector[K] log_theta = log(theta);  
}
model {
  mu ~ multi_normal(rep_vector(0.0,2), diag_matrix(rep_vector(1.0,2)));
  for (n in 1:N) {
    vector[K] lps = log_theta;
    for (k in 1:K)
      lps[k] += multi_normal_lpdf(y[n] | mu[k], diag_matrix(rep_vector(1.0,2)));
    target += beta*log_sum_exp(lps);
  }
}
generated quantities{
  vector[N] log_lik;
  for (n in 1:N) {
    vector[K] lps = log_theta;
    for (k in 1:K)
      lps[k] += multi_normal_lpdf(y[n] | mu[k], diag_matrix(rep_vector(1.0,2)));
    log_lik[n] = log_sum_exp(lps);
  }
}
\end{lstlisting}

\end{document}